# Object Detection in Optical Remote Sensing Images: A Survey and A New Benchmark

Ke Li[1], Gang Wan[1], Gong Cheng[2]*, Liqiu Meng[3], Junwei Han[2]*

[1] *Zhengzhou Institute of Surveying and Mapping, Zhengzhou 450052, China*
[2]*School of Automation, Northwestern Polytechnical University, Xi'an 710072, China*
[3]*Department of Cartography, Technical University of Munich, Arcisstr.21 80333 Munich, Germany*

**Abstract**: Substantial efforts have been devoted more recently to presenting various methods for object detection in optical remote sensing images. However, the current survey of datasets and deep learning based methods for object detection in optical remote sensing images is not adequate. Moreover, most of the existing datasets have some shortcomings, for example, the numbers of images and object categories are small scale, and the image diversity and variations are insufficient. These limitations greatly affect the development of deep learning based object detection methods. In the paper, we provide a comprehensive review of the recent deep learning based object detection progress in both the computer vision and earth observation communities. Then, we propose a large-scale, publicly available benchmark for object DetectIon in Optical Remote sensing images, which we name as DIOR. The dataset contains 23463 images and 192472 instances, covering 20 object classes. The proposed DIOR dataset 1) is large-scale on the object categories, on the object instance number, and on the total image number; 2) has a large range of object size variations, not only in terms of spatial resolutions, but also in the aspect of inter- and intra-class size variability across objects; 3) holds big variations as the images are obtained with different imaging conditions, weathers, seasons, and image quality; and 4) has high inter-class similarity and intra-class diversity. The proposed benchmark can help the researchers to develop and validate their data-driven methods. Finally, we evaluate several state-of-the-art approaches on our DIOR dataset to establish a baseline for future research.

## 1. Introduction

The rapid development of remote sensing techniques has significantly increased the quantity and quality of remote sensing images available to characterize various objects on the earth surface, such as airports, airplanes, buildings, etc. This naturally brings a strong requirement for intelligent earth observation through automatic analysis and understanding of satellite or aerial images. Object detection plays a crucial role in image interpretation and also is very important for a wide scope of applications, such as intelligent monitoring, urban planning, precision agriculture, and geographic information system (GIS) updating. Driven by this requirement, significant efforts have been made in the past few years to develop a variety of methods for object detection in optical remote sensing images (Aksoy, 2014; Bai et al., 2014; Cheng et al., 2013a; Cheng and Han, 2016; Cheng et al., 2013b; Cheng et al., 2014; Cheng et al., 2019; Cheng et al., 2016a; Das et al., 2011; Han et al., 2015; Han et al., 2014; Li et al., 2018; Long et al., 2017; Tang et al., 2017b; Yang et al., 2017; Zhang et al., 2016; Zhang et al., 2017; Zhou et al., 2016).

More recently, deep learning based algorithms have been dominating the top accuracy benchmarks for various visual recognition tasks (Chen et al., 2018; Cheng et al., 2018a; Clément et al., 2013; Ding et al., 2017; Hinton et al., 2012; Hou et al., 2017; Krizhevsky et al., 2012; Mikolov et al., 2012; Tian et al., 2017; Tompson et al., 2014; Wei et al., 2018) because of their powerful feature representation capabilities. Benefiting from this and some publicly available natural image datasets such as Microsoft Common Objects in Context (MSCOCO) (Lin et al., 2014) and PASCAL Visual Object Classes (VOC) (Everingham et al., 2010), a number of deep learning based object detection approaches

---

have achieved great success in natural scene images (Agarwal et al., 2018; Dai et al., 2016; Girshick, 2015; Girshick et al., 2014; Han et al., 2018; Liu et al., 2018a; Liu et al., 2016a; Redmon et al., 2016; Redmon and Farhadi, 2017; Ren et al., 2017).

However, despite the significant success achieved in natural images, it is difficult to straight-forward transfer deep learning based object detection methods to optical remote sensing images. As we know, high-quality and large-scale datasets are greatly important for training deep learning based object detection methods. However, the difference between remote sensing images and natural scene images is significant. As shown in Fig. 1, the remote sensing images generally capture the roof information of the geospatial objects, whereas the natural scene images usually capture the profile information of the objects. Therefore, it is not surprising that the object detectors learned from natural scene images are not easily transferable to remote sensing images. Although several popular object detection datasets, such as NWPU VHR-10 (Cheng et al., 2016a), UCAS-AOD (Zhu et al., 2015a), COWC (Mundhenk et al., 2016), and DOTA (Xia et al., 2018), are proposed in the earth observation community, they are still far from satisfying the requirements of deep learning algorithms.

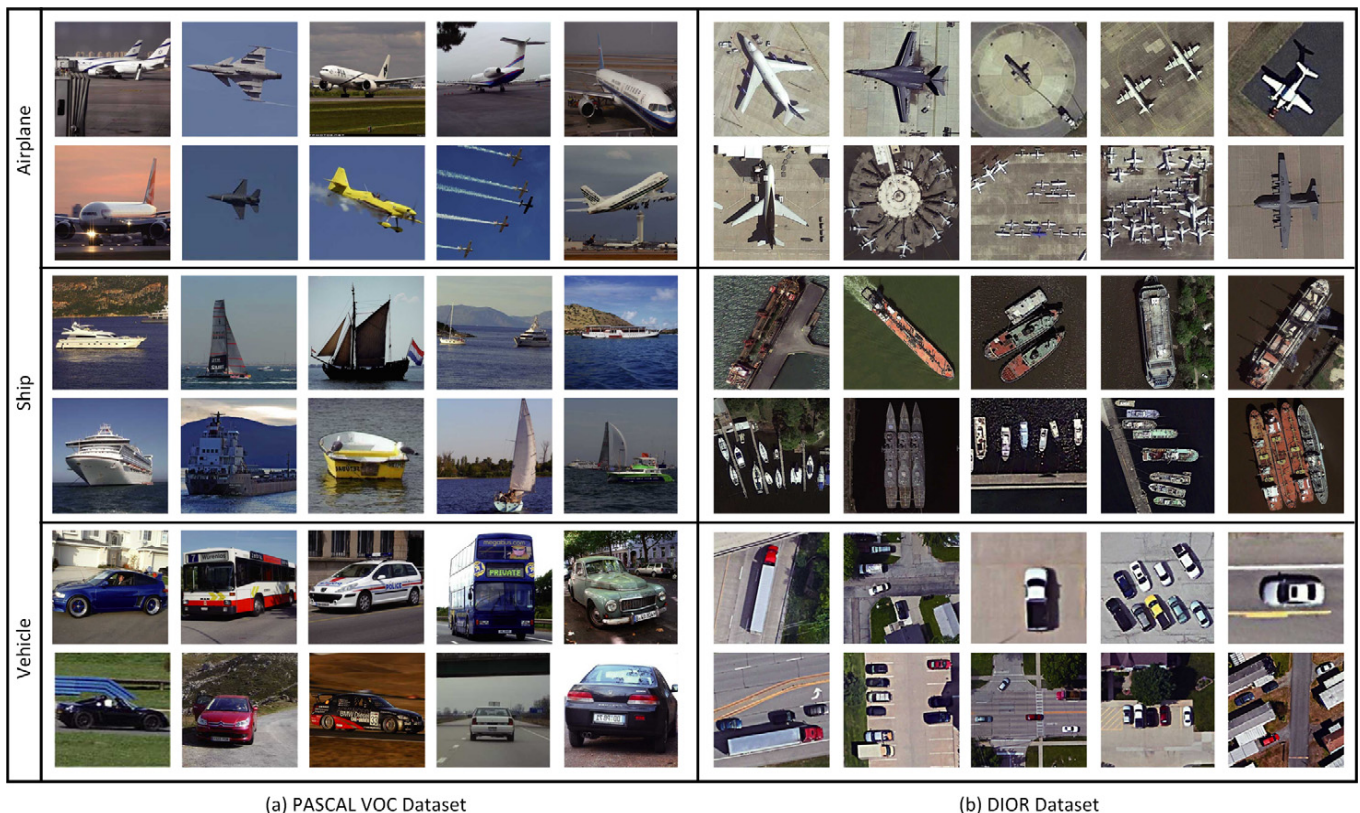

**Fig. 1**. Some examples, taken from (a) the PASCAL VOC dataset and (b) the proposed DIOR dataset, demonstrate the difference between natural scene images and remote sensing images.

To date, significant efforts (Cheng and Han, 2016; Cheng et al., 2016a; Das et al., 2011; Han et al., 2015; Li et al., 2018; Razakarivony and Jurie, 2015; Tang et al., 2017b; Xia et al., 2018; Yokoya and Iwasaki, 2015; Zhang et al., 2016; Zhu et al., 2017) have been made for object detection in remote sensing images. However, the current survey of the literatures concerning the datasets and deep learning based object detection methods is still not adequate. Moreover, most of the existing publicly available datasets have some shortcomings, for example, the numbers of images and object categories are small scale, and the image diversity and variations are also insufficient. These limitations greatly block the development of deep learning based object detection methods.

In order to address the aforementioned problems, we attempt to comprehensively review the recent progress of deep learning based object detection methods. Then, we propose a large-scale, publicly available benchmark for object DetectIon in Optical Remote sensing images, which we name as DIOR. Our proposed dataset consists of 23463 images covered by 20 object categories and each category contains about 1200 images. We highlight four key characteristics of the proposed DIOR dataset when comparing it with other existing object detection datasets. First, the numbers of total images, object categories, and object instances are large-scale. Second, the objects have a large range of size variations, not only in terms of spatial resolutions, but also in the aspect of inter- and intra-class size variability across objects. Third, our dataset holds large variations because the images are obtained with different imaging conditions, weathers, seasons, and image quality. Fourth, it possesses high inter-class similarity and intra-class diversity. Fig. 2 shows some example images and their annotations from our proposed DIOR dataset.

Our main contributions are summarized as follows:

**1) Comprehensive survey of deep learning based object detection progress**. We review the recent progress of existing datasets and representative deep learning based methods for object detection in both the computer vision and earth observation communities, which covers more than 110 papers.

**2) Creation of large-scale benchmark dataset**. This paper proposes a large-scale, publicly available dataset for object detection in optical remote sensing images. The proposed DIOR dataset, to our best knowledge, is the largest scale on both the number of object categories and the total number of images. The dataset enables the community to validate and develop data-driven object detection methods.

**3) Performance benchmarking on the proposed DIOR dataset.** We benchmark several representative deep learning based object detection methods on our DIOR dataset to provide an overview of the state-of-the-art performance for future research work.

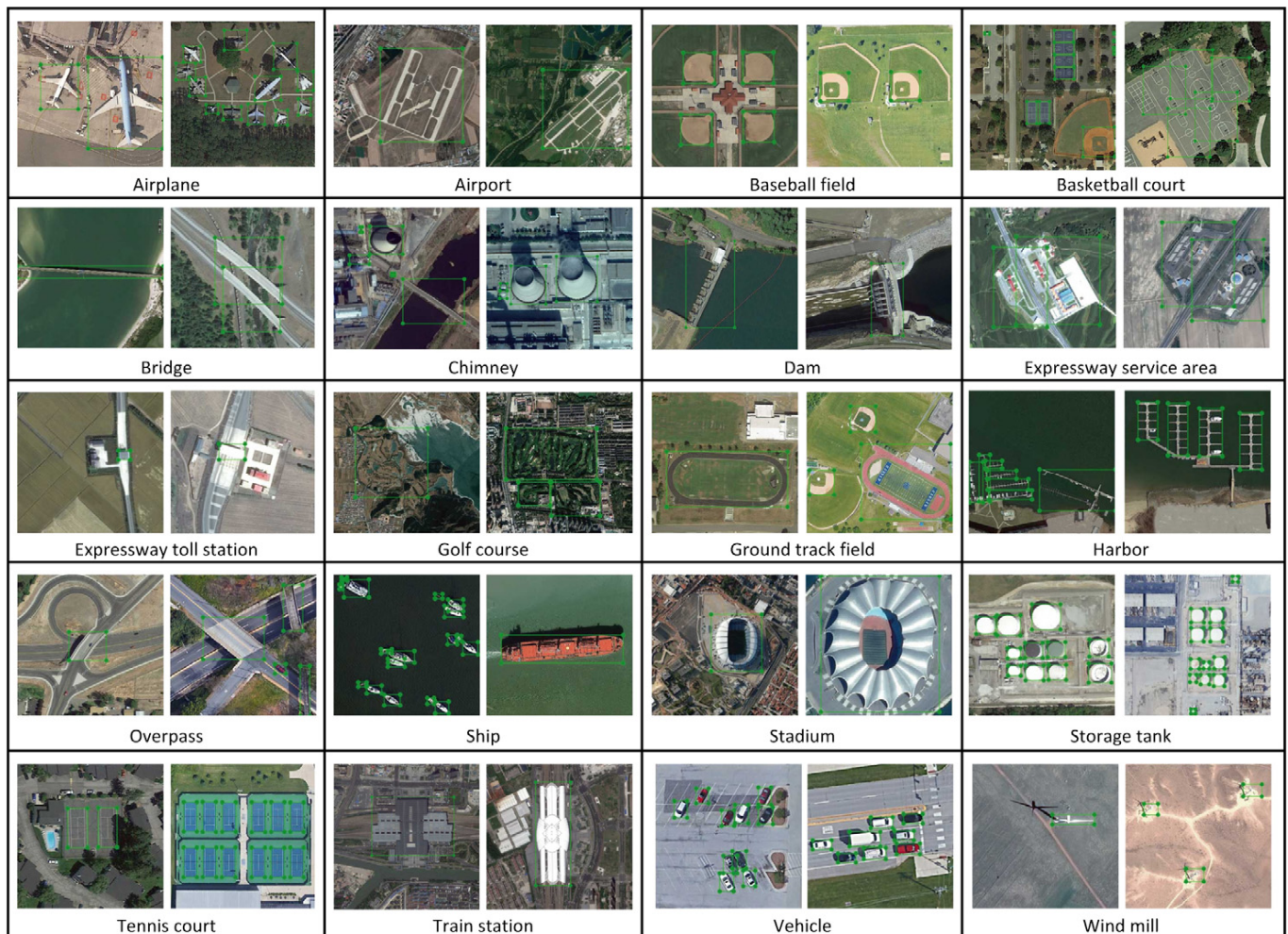

**Fig. 2**. Example images taken from the proposed DIOR dataset, which were obtained with different imaging conditions, weathers, seasons, and image quality.

The remainder of this paper is organized as follows. Sections 2-3 review the recent object detection progresses of benchmark datasets and deep learning methods in computer vision and earth observation community, respectively. Section 4 describes the proposed DIOR dataset in detail. Section 5 benchmarks several representative deep learning based object detection methods on the proposed dataset. Finally, Section 6 concludes this paper.

## 2. Review on Object Detection in Computer Vision Community

With the emergence of a variety of deep learning models, especially Convolutional Neural Networks (CNN), and their great success on image classification (He et al., 2016; Krizhevsky et al., 2012; Luan et al., 2018; Simonyan and Zisserman, 2015; Szegedy et al., 2015), numerous deep learning based object detection frameworks have been proposed in the computer vision community. Therefore, we will first provide a systematic survey of the references about the datasets as well as deep learning based methods for the task of object detection in natural scene images.

### 2.1 Object Detection Datasets of Natural Scene Images

Large-scale and high-quality datasets are very important for boosting object detection performance, especially for deep learning based methods. The PASCAL VOC (Everingham et al., 2010), MSCOCO (Lin et al., 2014), and ImageNet object detection dataset (Deng et al., 2009) are three widely used datasets for object detection in natural scene images. These datasets are briefly reviewed as follows.

*1) PASCAL VOC Dataset.* The PASCAL VOC 2007 (Everingham et al., 2010) and VOC 2012 (Everingham et al., 2015) are two most-used datasets for natural scene image object detection. Both of them contain 20 object classes, but with different image numbers. Specifically, the PASCAL VOC 2007 dataset contains a total of 9963 images with 5011 images for training and 4952 images for testing. The PASCAL VOC 2012 dataset extends the PASCAL VOC 2007 dataset, resulting in a larger scale dataset that consists of 11540 images for training and 10991 images for testing.

*2) MSCOCO Dataset.* The MSCOCO dataset was proposed by Microsoft in 2014 (Lin et al., 2014). The scale of MSCOCO dataset is much larger than the PASCAL VOC dataset on both the number of object categories and object instances. Specifically, the dataset consists of more than 200000 images covered by 80 object categories. The dataset is further divided into three subsets: training set, validation set and testing set, which contain about 80k, 40k, and 80k images, respectively.

*3) ImageNet Object Detection Dataset.* This dataset was released in 2013 (Deng et al., 2009), which has the most object categories and the largest number of images among all object detection datasets. Specifically, this dataset includes 200 object classes and more than 500000 images, with 456567 images for training, 20121 images for validation, and 40152 images for testing, respectively.

### 2.2 Deep Learning Based Object Detection Methods in Computer Vision Community

Recently, a number of deep learning based object detection methods have been proposed, which significantly improve the performance of object detection. Generally, the existing deep learning methods designed for object detection can be divided into two streams on the basis of whether or not generating region proposals. They are region proposal-based methods and regression-based methods.

*2.2.1 Region Proposal-based Methods*

In the past few years, region proposal-based object detection methods have achieved great success in natural scene images (Dai et al., 2016; Girshick, 2015; Girshick et al., 2014; He et al., 2017; He et al., 2014; Lin et al., 2017b; Ren et al., 2017). This kind of approaches divides the framework of object detection into two stages. The first stage focuses on generating a series of candidate region proposals that may contain objects. The second stage aims to classify the candidate region proposals obtained from the first stage into object classes or background and further fine-tune the coordinates of the bounding boxes.

The Region-based CNN (R-CNN) proposed by Girshick *et al.* (Girshick et al., 2014) is one of the most famous approaches in various region proposal-based methods. It is the representative work to adopt the CNN models to generate rich features for object detection, achieving breakthrough performance improvement in comparison with all previously works, which are mainly based on deformable part model (DPM) (Felzenszwalb et al., 2010). Briefly, R-CNN consists of three simple steps. First, it scans the input image for possible objects by using selective search method (Uijlings et al., 2013), generating about 2000 region proposals. Second, these region proposals are resized into a fixed size (e.g., 224×224) and the deep features of each region proposal are extracted by using a CNN model fine-tuned on the PASCAL VOC dataset. Finally, the features of each region proposal are fed into a set of class-specific support vector machines (SVMs) to label each region proposal as object or background and a linear regressor is used to refine the object localizations (if object exist).

While R-CNN surpasses previous object detection methods, the low efficiency is its main shortcoming due to the repeated computation of abundant region proposals. In order to obtain better detection efficiency and accuracy, some recent works, such as SPPnet (He et al., 2014) and Fast R-CNN (Girshick, 2015), were proposed for sharing the computation load of CNN feature extraction of all region proposals. Compared with R-CNN, Fast R-CNN and SPPnet perform feature extraction over the whole image with a region of interest (RoI) layer and a spatial pyramid pooling (SPP) layer, respectively, in which the CNN model runs over the entire image only once rather than thousands of times, thus they need less computation time than R-CNN.

Although SPPnet and Fast R-CNN work at faster speeds than R-CNN, they need obtaining region proposals in advance, which are usually generated with hand-engineering proposal detectors such as EdgeBox (Zitnick and Dollár, 2014) and selective search method (Uijlings et al., 2013). However, the handcrafted region proposal mechanism is a severe bottleneck of the entire object detection process. Thus, Faster R-CNN (Ren et al., 2017) was proposed in order to fix this problem. The main insight of Faster R-CNN is to adopt a fast module to generate region proposals instead of the slow selective search algorithm. Specifically, the Faster R-CNN framework consists of two modules. The first model is the region proposal network (RPN), which is a fully convolutional network used to generate region proposals. The second module is the Fast R-CNN object detector used for classifying the proposals which are generate with the first module. The core idea of the Faster R-CNN is to share the same convolutional layers for the RPN and Fast R-CNN detector up to their own fully connected layers. In this way, the image only needs to pass through the CNN once to generate region proposals and their corresponding features. More importantly, thanks to the sharing of convolutional layers, it is possible to use a very deep CNN model to generate higher-quality region proposals than traditional region proposal generation methods.

In addition, some researchers further extend the work of Faster R-CNN for better performance. For example, Mask R-CNN (He et al., 2017) builds on Faster R-CNN and adds an additional branch to predict an object mask in parallel with the existing branch for bounding box detection. Thus, Mask R-CNN can accurately recognize objects and simultaneously generate high-quality segmentation masks for each object instance. In order to further speed up object detection of Faster R-CNN, region-based fully convolutional network (R-FCN) (Dai et al., 2016) was proposed. It uses a position-sensitive region of interest (RoI) pooling layer to aggregate the outputs of the last convolutional layer and produce scores for each RoI. In contrast to Faster R-CNN that applies a costly per-region sub-network hundreds of times, R-FCN shares almost all computation load on the entire image, resulting in 2.5-20× faster than Faster R-CNN. Besides, Li *et al.* (Li et al., 2017) proposed a Light Head R-CNN to further speed up the detection speed of R-FCN (Dai et al., 2016) by making the head of the detection network as light as possible. Also, Singh et al. proposed a novel detector, called R-FCN-3000 (Singh et al., 2018a), towards large-scale real-time object detection for 3000 object classes. This approach is a modification of R-FCN (Dai et al., 2016) used to learn shared filters to perform localization across different object classes.

In 2017, a feature pyramid network (FPN) (Lin et al., 2017b) was proposed by building feature pyramids inside CNNs, which shows significant improvement as a generic feature extractor for object detection with the frameworks of Faster R-CNN (Ren et al., 2017) and Mask R-CNN (He et al., 2017). Also, a path aggregation network (PANet) (Liu et al., 2018b) was proposed to boost the entire feature hierarchy with accurate localization information in lower layers by bottom-up path augmentation, which can significantly shorten the information path between lower layers and topmost feature.

More recently, Singh et al. proposed two advanced and effective data argumentation methods for object detection, including Scale Normalization for Image Pyramids (SNIP) (Singh and Davis, 2018) and SNIP with Efficient Resampling (SNIPER) (Singh et al., 2018b). These two methods present detailed analysis of different techniques for detecting and recognizing objects under extreme scale variation. To be specific, SNIP (Singh and Davis, 2018) is a novel training paradigm that builds image pyramids at both of training and detection stages and only selectively back-propagates the gradients of objects of different sizes as a function of the image scale. Thus, it significantly benefits from reducing scale-variation during training but without reducing training samples. SNIPER (Singh et al., 2018b) is an efficient multi-scale training approach proposed to adaptively generate training samples from multiple scales of an image pyramid, conditioned on the image content. Under the same conditions, SNIPER performs as well as SNIP while reducing the number of pixels processed by a factor of 3 during training. Here, it should be pointed out that SNIP (Singh and Davis, 2018) and SNIPER (Singh et al., 2018b) are generic and thus can be broadly applied to many detectors, such as Faster R-CNN (Ren et al., 2017), Mask R-CNN (He et al., 2017), R-FCN (Dai et al., 2016), deformable R-FCN (Dai et al., 2017), and so on.

*2.2.2 Regression-based Methods*

This kind of methods uses one-stage object detectors for object instance prediction, thus simplifying detection as a regression problem. Compared with region proposal-based methods, regression-based methods are much simpler and more efficient, because there is no need to produce candidate region proposals and the subsequent feature re-sampling stages. OverFeat (Sermanet et al., 2014) is the first regression-based object detector based on deep networks using sliding-window paradigm. More recently, You Only Look Once (YOLO) (Redmon et al., 2016; Redmon and Farhadi, 2017, 2018), Single Shot multibox Detector (SSD) (Fu et al., 2017; Liu et al., 2016a), and RetinaNet (Lin et al., 2017c) have renewed the performance in regression-based methods.

YOLO (Redmon et al., 2016) is a representative regression-based object detection method. It adopts a single CNN backbone to directly predict bounding boxes and class probabilities from the entire images in one evaluation. It works as follows. Given an input image, it is firstly divided into $S \times S$ grids. If the center of an object falls into a grid cell, that grid is responsible for the detection of that object. Then, each grid cell predicts $B$ bounding boxes together with their confidence scores and $C$ class probabilities. YOLO achieves object detection in real-time by reframing it as a single regression problem. However, it still struggles to precisely localize some objects, especially small-sized objects.

In order to improve both the speed and accuracy, SSD (Liu et al., 2016a) was proposed. Specifically, the output space of bounding boxes is discretized into a set of default boxes over different scales and aspect ratios per feature map location. At prediction process, the confidence scores for the presence of each object class in each default box are generated based on the SSD model and the adjustments to the box are also produced to better match the object shape. Furthermore, in order to address the problem of object size variations, SSD combines the predictions obtained from multiple feature maps with different resolutions. Compared with YOLO (Redmon et al., 2016), SSD achieved better performance for detecting and locating small-sized objects via introducing default boxes mechanism and multi-scale feature maps. Another interesting work is RetinaNet detector (Lin et al., 2017c), which is essentially a feature pyramid network with the traditional cross-entropy loss being replaced by a new Focal loss (Lin et al., 2017c), and thereby increasing the accuracy significantly.

The insight of YOLOv2 model (Redmon and Farhadi, 2017) is to improve object detection accuracy while still being an efficient object detector. To this end, it proposes various improvements to the original YOLO method. For example, in order to prevent over-fitting without using dropout, YOLOv2 adds the batch normalization on all of the convolutional layers. It accepts higher-resolution images as input by adjusting the input image size from 224×224 (YOLO) to 448×448 (YOLOv2), thus the objects with smaller sizes can be detected effectively. Additionally, YOLOv2 removes the fully-connected layers from the original YOLO detector and predicts bounding boxes based on anchor boxes, which shares the similar idea with SSD (Liu et al., 2016a).

More recently, YOLOv3 model (Redmon and Farhadi, 2018), which has similar performance but is faster than YOLOv2, SSD, and RetinaNet, was proposed. YOLOv3 adheres to YOLOv2's mechanism. To be specific, the bounding boxes are predicted using dimension clusters as anchor boxes. Then, independent logistic classifiers instead of softmax classifier are adopted to output an object score for each bounding box. Sharing a similar concept with FPN (Lin et al., 2017b), the bounding boxes are predicted at three different scales through extracting features from these scales. YOLOv3 uses a new backbone network, named Darketnet-53, for performing feature extraction. It has 53 convolutional layers, and is a newfangled residual network. Due to the introduction of Darketnet-53 and multi-scale feature maps, YOLOv3 achieves great speed improvement and also improves detection accuracy of small-sized objects when compared with the original YOLO or YOLOv2.

In addition, Law and Deng proposed CornerNet (Law and Deng, 2018), a new and effective object detection paradigm that detects object bounding boxes as pairs of corners (i.e., the top-left corner and the bottom-right corner) by using a single CNN. By detecting objects as paired corners, CornerNet eliminates the need for designing a set of anchor boxes widely used in regression-based object detectors. This work also introduces corner pooling, a new type of pooling layer that helps the network better localize corners.

In general, region proposal-based object detection methods have better accuracies than regression-based algorithms, while regression-based algorithms have advantages in speed (Lin et al., 2017c). It is generally accepted that CNN framework plays a crucial role in object detection task. CNN architectures serve as network backbones used in various object detection frameworks. Some representative CNN model architectures include AlexNet (Krizhevsky et al., 2012), ZFNet (Zeiler and Fergus, 2014), VGGNet (Simonyan and Zisserman, 2015), GoogLeNet (Szegedy et al., 2015), Inception series (Ioffe and Szegedy, 2015; Szegedy et al., 2017; Szegedy et al., 2016), ResNet (He et al., 2016), DenseNet (Huang et al., 2017) and SENet (Hu et al., 2018). Also, some researches have been widely explored to further improve the performance of deep learning based methods for object detection, such as feature enhancement (Cai et al., 2016; Cheng et al., 2019; Cheng et al., 2016b; Kong et al., 2016; Liu et al., 2017b), hard negative mining (Lin et al., 2017c; Shrivastava et al., 2016), contextual information fusion (Bell et al., 2016; Gidaris and Komodakis, 2015; Shrivastava and Gupta, 2016; Zhu et al., 2015b), modeling object deformations (Mordan et al., 2018; Ouyang et al., 2017; Xu et al., 2017), and so on.

## 3. Review on Object Detection in Earth Observation Community

In the past years, numerous object detection approaches have been explored to detect various geospatial objects in the earth observation community. Cheng *et al* (Cheng and Han, 2016) provide a comprehensive review in 2016 on object detection algorithms in optical remote sensing images. However, the work of (Cheng and Han, 2016) does not review various deep learning based object detection methods. Different from several previously published surveys, we focus on reviewing the literatures about datasets and deep learning based approaches for object detection in the earth observation community.

### 3.1 Object Detection Datasets of Optical Remote Sensing Images

During the last decades, several different research groups have released their publicly available earth observation image datasets for object detection (see Table 1). These datasets will be briefly reviewed as follows.

**Table 1**
Comparison between the proposed DIOR dataset and nine publicly available object detection datasets in earth observation community.

| Datasets | # Categories | # Images | # Instances | Image width | Annotation way | Year |
| --- | --- | --- | --- | --- | --- | --- |
| TAS | 1 | 30 | 1319 | 792 | horizontal bounding box | 2008 |
| SZTAKI-INRIA | 1 | 9 | 665 | ~800 | oriented bounding box | 2012 |
| NWPU VHR-10 | 10 | 800 | 3775 | ~1000 | horizontal bounding box | 2014 |
| VEDAI | 9 | 1210 | 3640 | 1024 | oriented bounding box | 2015 |
| UCAS-AOD | 2 | 910 | 6029 | 1280 | horizontal bounding box | 2015 |
| DLR 3K Vehicle | 2 | 20 | 14235 | 5616 | oriented bounding box | 2015 |
| HRSC2016 | 1 | 1070 | 2976 | ~1000 | oriented bounding box | 2016 |
| RSOD | 4 | 976 | 6950 | ~1000 | horizontal bounding box | 2017 |
| DOTA | 15 | 2806 | 188282 | 800-4000 | oriented bounding box | 2017 |
| DIOR (ours) | 20 | 23463 | 192472 | 800 | horizontal bounding box | 2018 |

*1) TAS:* The TAS dataset (Heitz and Koller, 2008) is designed for car detection in aerial images. It contains a total of 30 images and 1319 manually annotated cars with arbitrary orientations. These images have relatively low spatial resolution and a lot of shadows caused by buildings and trees.

*2) SZTAKI-INRIA:* The SZTAKI-INRIA dataset (Benedek et al., 2011) is created for benchmarking various building

detection methods. It consists of 665 buildings, manually annotated with oriented bounding boxes, distributed throughout nine remote sensing images derived from Manchester (U.K.), Szada and Budapest (Hungary), Cot d'Azur and Normandy (France), and Bodensee (Germany). All of the images contain only red (R), green (G), and blue (B) three channels. Among them, two images (Szada and Budapest) are aerial images and the rest seven images are satellite images from QuickBird, IKONOS, and Google Earth.

*3) NWPU VHR-10:* The NWPU VHR-10 dataset (Cheng and Han, 2016; Cheng et al., 2016a) has 10 geospatial object classes including airplane, baseball diamond, basketball court, bridge, harbor, ground track field, ship, storage tank, tennis court, and vehicle. It consists of 715 RGB images and 85 pan-sharpened color infrared images. To be specific, the 715 RGB images are collected from Google Earth and their spatial resolutions vary from 0.5m to 2m. The 85 pan-sharpened infrared images, with a spatial resolution of 0.08m, are obtained from Vaihingen data (Cramer, 2010). This dataset contains a total of 3775 object instances which are manually annotated with horizontal bounding boxes, including 757 airplanes, 390 baseball diamonds, 159 basketball courts, 124 bridges, 224 harbors, 163 ground track fields, 302 ships, 655 storage tanks, 524 tennis courts, and 477 vehicles. This dataset has been widely used in the earth observation community (Cheng et al., 2014; Cheng et al., 2018b; Farooq et al., 2017; Guo et al., 2018; Han et al., 2017a; Li et al., 2018; Yang et al., 2018b; Yang et al., 2017; Zhong et al., 2018).

*4) VEDAI:* The VEDAI (Razakarivony and Jurie, 2015) dataset is released for the task of multi-class vehicle detection in aerial images. It consists of 3640 vehicle instances covered by nine classes including boat, car, camping car, plane, pick-up, tractor, truck, van, and the other category. This dataset contains totally 1210 1024×1024 aerial images acquired from Utah AGRC (http://gis.utah.gov/), with a spatial resolution of 12.5 cm. The images in the dataset were captured during spring 2012 and each image has four uncompressed color channels including three RGB color channels and one near infrared channel.

*5) UCAS-AOD:* The UCAS-AOD dataset (Zhu et al., 2015a) is designed for airplane and vehicle detection. Specifically, the airplane dataset consists of 600 images with 3210 airplanes and the vehicle dataset consists of 310 images with 2819 vehicles. All the images are carefully selected so that the object orientations in the dataset distribute evenly.

*6) DLR 3K Vehicle:* The DLR 3K Vehicle dataset (Liu and Mattyus, 2015) is another dataset designed for vehicle detection. It contains 20 5616×3744 aerial images, with a spatial resolution of 13 cm. They are captured at a height of 1000 meters above the ground using the DLR 3K camera system (a near real time airborne digital monitoring system) over the area of Munich, Germany. There are 14235 vehicles that are manually labeled by using oriented bounding boxes in the images.

*7) HRSC2016:* The HRSC2016 dataset (Liu et al., 2016b) contains 1070 images and a total of 2976 ships collected from Google Earth used for ship detection. The image sizes change from 300×300 to 1500×900, and most of them are about 1000×600. These images are collected with large variations of rotation, scale, position, shape, and appearance.

*8) RSOD:* The RSOD dataset (Xiao et al., 2015) contains 976 images downloaded from Google Earth and Tianditu, and the spatial resolutions of these images range from 0.3m to 3m. It consists of totally 6950 object instances, covered by four object classes, including 1586 oil tanks, 4993 airplanes, 180 overpasses, and 191 playgrounds.

*9) DOTA:* The DOTA (Xia et al., 2018) is a new large-scale geospatial object detection dataset, which consists of 15 different object categories: baseball diamond, basketball court, bridge, harbor, helicopter, ground track field, large vehicle, plane, ship, small vehicle, soccer ball field, storage tank, swimming pool, tennis court, and roundabout. This dataset contains a total of 2806 aerial images obtained from different sensors and platforms with multiple resolutions. There are 188282 object instances labeled by an oriented bounding box. The sizes of images range from about 800×800 to 4000×4000 pixels. Each image contains multiple objects of different scales, orientations and shapes. To date, this dataset is the most challenging.

**3.2 Deep Learning Based Object Detection Methods in Earth Observation Community**

Inspired by the great success of deep learning-based object detection methods in computer vision community, extensive studies have been devoted recently to object detection in optical remote sensing images. Different from object detection in natural scene mages, most of the studies use region proposal-based methods to detect multi-class objects in the earth observation community. We therefore no longer distinguish them between region proposal-based methods or regression-based methods in the earth observation community. Here, we mainly review some representative methods.

Driven by the excellent performance of R-CNN for natural scene image object detection, a number of earth observation researchers adopt R-CNN pipeline to detect various geospatial objects in remote sensing images (Cheng et al., 2016a; Long et al., 2017; Salberg, 2015; Ševo and Avramović, 2017). For instance, Cheng *et al*. (Cheng et al., 2016a) proposed to learn a rotation-invariant CNN (RICNN) model in R-CNN framework used for multi-class geospatial object detection. The RICNN is achieved by adding a new rotation-invariant layer to the off-the-shelf CNN model such as AlexNet (Krizhevsky et al., 2012). In order to further boost the state-of-the-arts of object detection, (Cheng et al., 2019) proposed a new method to train rotation-invariant and Fisher discriminative CNN (RIFD-CNN) model by imposing a rotation-invariant regularizer and a Fisher discrimination regularizer on the CNN features. To achieve accurate localization of geospatial objects in high-resolution earth observation images, Long *et al*. (Long et al., 2017) presented an unsupervised score-based bounding box regression (USB-BBR) method based on R-CNN framework.

Although the aforementioned methods have achieved good performance in the earth observation community, they are still time-consuming because these approaches depend on human-designed object proposal generation methods which domain most of the running time of an object detection system. In addition, the quality of region proposals generated based on hand-engineered low-level features is not good, thus thereby degenerating object detection performance.

In order to further enhance the detection accuracy and speed, a few research works extend the framework of Faster R-CNN to the earth observation community (Deng et al., 2017; Guo et al., 2018; Han et al., 2017b; Li et al., 2018; Tang et al., 2017b; Xu et al., 2017; Yang et al., 2018a; Yang et al., 2017; Yao et al., 2017; Zhong et al., 2018). For instance, Li *et al*. (Li et al., 2018) presented a rotation-insensitive RPN by introducing multi-angle anchors into the existing RPN based on Faster R-CNN pipeline, which can effectively handle the problem of geospatial object rotation variations. Furthermore, in order to tackle the problem of appearance ambiguity, a double-channel feature combination network is designed to learn local and contextual properties. Zhong *et al*. (Zhong et al., 2018) utilized a position-sensitive balancing (PSB) method to enhance the quality of generated region proposal. In the proposed PSB framework, a fully convolutional network (FCN) (Long et al., 2015) was introduced, based on the residual network (He et al., 2016), to address the dilemma between translation-variance in object detection and translation-invariance in image classification. Xu *et al*. (Xu et al., 2017) presented a deformable CNN to model the geometric variations of objects. In (Xu et al., 2017), non-maximum suppression constrained by aspect ratio was developed to reduce the increase of false region proposals. Aiming at vehicle detection, Tang *et al*. (Tang et al., 2017b) proposed a hyper region proposal network (HRPN) to find vehicle-like regions and use hard negative mining to further improve the detection accuracy.

Although adapting region proposal-based methods, such as R-CNN, Faster R-CNN, and their variants, to detect geospatial objects in earth observation images shows greatly promising performance, remarkable efforts have been made to explore different deep learning based methods (Lin et al., 2017a; Liu et al., 2017a; Liu et al., 2018c; Tang et al., 2017a; Yu et al., 2015; Zou and Shi, 2016), which do not follow the pipeline of region proposal-based approaches to detect objects in remote sensing images. For example, Yu *et al*. (Yu et al., 2015) presented a rotation-invariant method to detect geospatial objects, in which super-pixel segmentation strategy is firstly used to produce local patches, then, deep Boltzmann machines are adopted to construct high-level feature representations of local patches, and finally a set of multi-scale Hough forests is built to cast rotation-invariant votes to locate object centroids. Zou *et al*. (Zou and Shi, 2016) used a singular value decompensation network to obtain ship-like regions and adopt feature pooling operation and a linear SVM classifier to verify each ship candidate for the task of ship detection. Although this detection framework is interesting, the training process is still clumsy and slow.

More recently, in order to achieve real-time object detection, a few studies attempt to transfer regression-based detection methods developed for natural scene images to remote sensing images. For instance, sharing the similar idea with SSD, Tang *et al*. (Tang et al., 2017a) used a regression-based object detector to detect vehicle targets. Specifically, the detection bounding boxes are generated by adopting a set of default boxes with different scales per feature map location. Moreover, for each default box the offsets are predicted to better fit the object shape. Liu *et al*. (Liu et al., 2017a) replaced the traditional bounding box with rotatable bounding box (RBox) embedded into SSD framework (Liu et al., 2016a), thus being rotation invariant due to its ability of estimating the orientation angles of objects. Liu *et al*. (Liu et al., 2018c) designed a framework for detecting arbitrary-oriented ships. This model can directly predict rotated/oriented bounding boxes by using YOLOv2 architecture as the fundamental network. In addition, hard example mining (Tang et al., 2017a; Tang et al., 2017b), multi-feature fusion (Zhong et al., 2017), transfer learning (Han et al., 2017b), non-maximum suppression (Xu et al., 2017), etc., are often used in geospatial object detection to further boost the performance of deep learning based approaches.

Although most of the existing deep learning based methods have demonstrated significant achievement on the task of object detection in the earth observation community, they are transferred from the methods (e.g., R-CNN, Faster R-CNN, SSD, etc.) designed for natural scene images. In fact, as we have pointed out above, earth observation images significantly differ from natural scene images is significant, especially in the aspects of rotation, scale variation and the complex and cluttered background. Although the existing methods partially addressed these issues through introducing prior knowledge or designing proprietary models, the task of object detection in earth observation images is still an open problem deserved to further research.

## 4. Proposed DIOR Dataset

In the last few years, remarkable efforts have been made to release various object detection datasets (reviewed in Section 3.2) in the earth observation community. However, most of existing object detection datasets in earth observation domain shares some common shortcomings, for example, the number of images and the number of object categories are small scale, and the image diversity and object variations are insufficient. These limitations significantly affect the development of deep learning based object detection methods. In such a situation, creating a large-scale object detection dataset by using remote sensing images is highly desirable for the earth observation community. This motivates us to create a large-scale dataset named DIOR. It is publicly available[2] and can be used freely for object detection in optical remote sensing images.

### 4.1 Object Class Selection

Selecting appropriate geospatial object classes is the first step of constructing the dataset and is crucial for the dataset. In our work, we first investigated the object classes of all existing datasets (Benedek et al., 2011; Cheng and Han, 2016; Cheng et al., 2016a; Heitz and Koller, 2008; Liu and Mattyus, 2015; Liu et al., 2016b; Razakarivony and Jurie, 2015; Xia et al., 2018; Xiao et al., 2015; Zhu et al., 2015a) to obtain 10 object categories which are commonly used in both NWPU VHR-10 dataset and DOTA dataset. We then further extend the object categories of our dataset by searching the keywords of "object detection", "object recognition", "earth observation images", and "remote sensing images" on Google Scholar and Web of Science to carefully select other 10 object classes, according to whether a kind of objects is common or its value for real-world applications. For example, some traffic infrastructures that are common and play an important role in transportation analysis, such as train stations, expressway service areas, and airports, are selected mainly because of their values in real applications. In addition, most of the object categories in existing datasets are selected from the urban areas. Therefore, dam and wind mill, which are common in the suburb as well as important infrastructures, are also chosen to improve the variations and diversity of geospatial objects. In such a situation, a total of 20 object classes are selected to create the proposed DIOR dataset. These 20 object classes are airplane, airport, baseball field, basketball court, bridge, chimney, dam, expressway service area, expressway toll station, harbor, golf course, ground track field, overpass, ship, stadium, storage tank, tennis court, train station, vehicle, and wind mill.

### 4.2 Characteristics of Our Proposed DIOR Dataset

The DIOR dataset is one of the largest, most diverse, and publicly available object detection dataset in earth observation community. We use LabelMe (Russell et al., 2008), an open-source image annotation tool, to annotate object instances. Each object instance is manually labeled by a horizontal bounding box which is typically used for object annotation in remote sensing images and natural scene images. Fig. 3 reports the number of object instances per class. In the proposed DIOR dataset, the object classes of ship and vehicle have higher instance counts, while the classes of train station, expressway toll station and expressway service area have lower instance counts. The diversity of object size is more helpful for real-world tasks. As shown in Fig. 4, we achieve a good balance between small-sized instances and big-sized instances. In addition, the significant object size differences across different categories makes the detection task more challenging, because which requires that the detectors have to be flexible enough to simultaneously handle small-sized and large-sized objects.

---

[2] http://www.escience.cn/people/gongcheng/DIOR.html

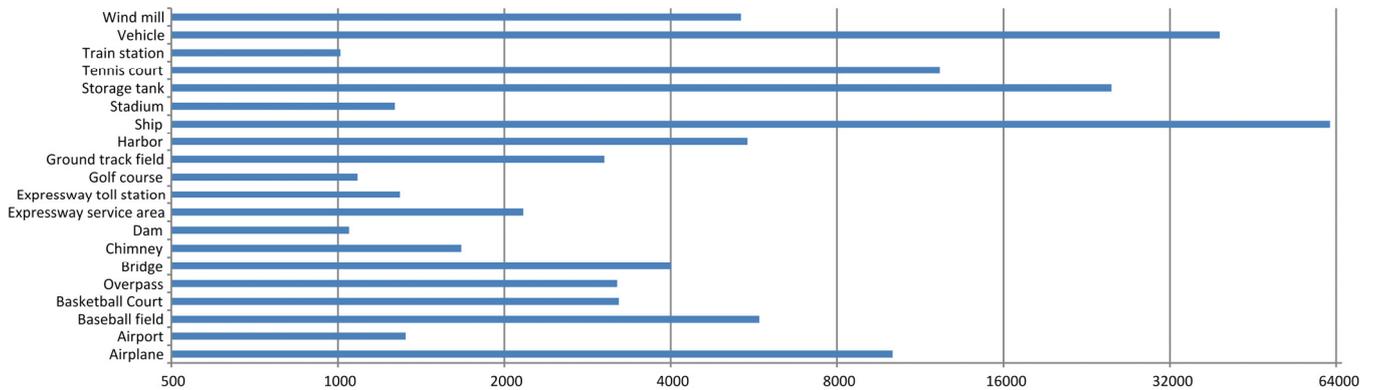

**Fig. 3**. Number of object instances per class.

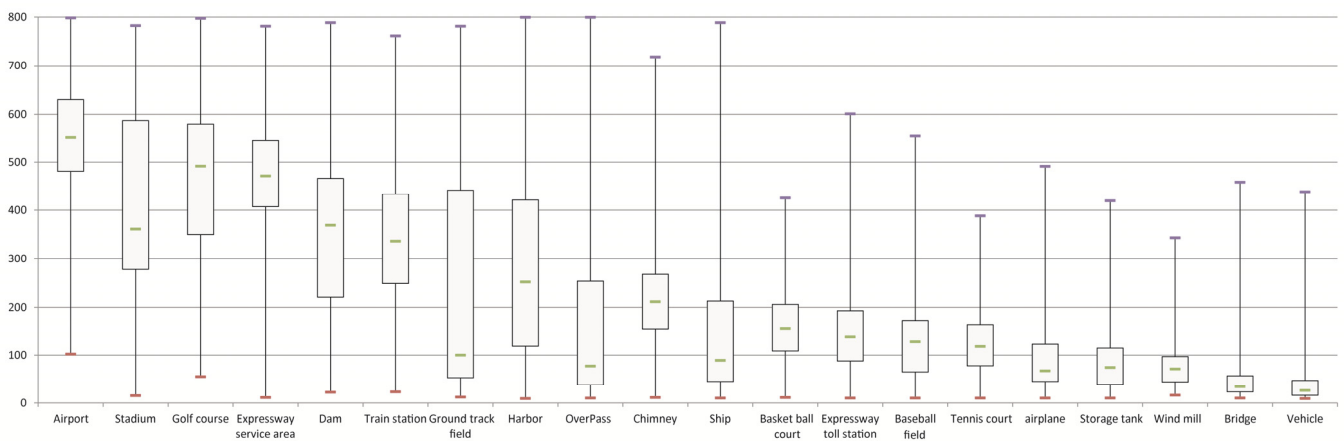

**Fig. 4**. Object size distribution per class.

Compared with existing object detection datasets including (Benedek et al., 2011; Cheng and Han, 2016; Cheng et al., 2016a; Heitz and Koller, 2008; Liu and Mattyus, 2015; Liu et al., 2016b; Razakarivony and Jurie, 2015; Tanner et al., 2009; Xia et al., 2018; Xiao et al., 2015; Zhu et al., 2015a), the proposed DIOR dataset has the following four remarkable characteristics.

*1) Large scale.* DIOR consists of 23463 optimal remote sensing images and 192472 object instances that are manually labeled with axis-aligned bounding boxes, covered by 20 common object categories. The size of images in the dataset is 800×800 pixels and the spatial resolutions range from 0.5m to 30m. Similar with most of the existing datasets, this dataset is also collected from Google Earth (Google Inc.), by the experts in the domain of earth observation interpretation.

Compared with all existing remote sensing image datasets designed for object detection, the proposed DIOR dataset, to our best knowledge, is the largest scale on both the number of images and the number of object categories. The release of this dataset will help the earth observation community to explore and evaluate a variety of deep learning based methods, thus thereby further improving the state of the arts.

*2) A large range of object size variations.* Spatial size variation is an important feature of geospatial objects. This is not only because of the spatial resolutions of sensors, but also due to between-class size variation (e.g., aircraft carriers *vs.* cars) and within-class size variation (e.g., aircraft carriers *vs.* hookers). There are a large range of size variations of object instances in the proposed DIOR dataset. To increase the size variations of objects, images with different spatial resolutions of objects are collected and the images which contain rich size variations coming from the same object category and different object categories are also collected in our dataset. As shown in Fig. 5 (a), "vehicle" and "ship" instances present different sizes. Besides, due to different spatial resolutions, the object sizes of "stadium" instances are also obviously different.

*3) Rich image variations.* A highly desired characteristic for any object detection system is its robustness to image variations. However, most of the existing datasets are lack of image variations totally or partially. For example, the widely used NWPU VHR-10 dataset only consists of 800 images, which is too small to possess much richer variations in various weathers, seasons, imaging conditions, scales, etc. On the contrary, the proposed DIOR dataset contains 23463 remote sensing images covered more than 80 countries. Moreover, these images are carefully collected under different weathers, seasons, imaging conditions, and image quality (see Fig. 5 (b)). Thus, our proposed DIOR dataset holds richer variations in viewpoint, translation, illumination, background, object pose and appearance, occlusion, etc., for each object class.

*4) High inter-class similarity and intra-class diversity.* Another important characteristic of our proposed dataset is that it has high inter-class similarity and intra-class diversity, thus making it much challenging. To obtain big inter-class similarity, we add some fine-grained object classes with high semantic overlapping, such as "bridge" *vs.* "overpass", "bridge" *vs.* "dam", "ground track field" *vs.* "stadium", "tennis court" *vs.* "basketball court", and so on. To increase intra-class diversity, all kinds of factors, such as different object colors, shapes and scales, are taken into account when collecting images. As shown in Fig. 5 (c), "chimney" instances present different shapes, and "dam" and "bridge" instances have very similar appearances.

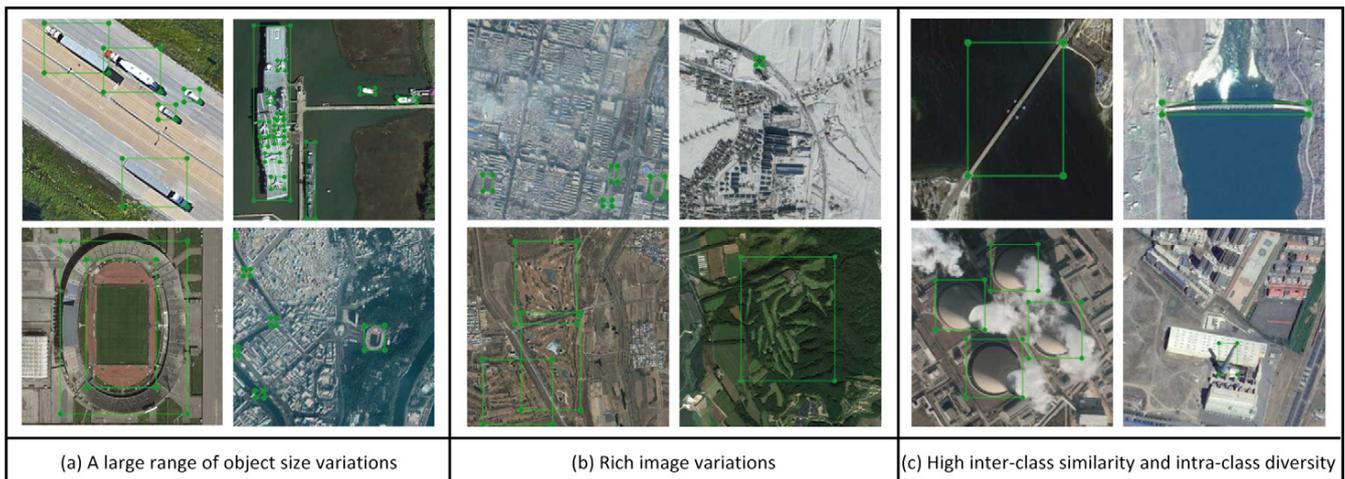

**Fig. 5**. Characteristics of our proposed DIOR dataset.

## 5. Benchmarking Representative Methods

This section focuses on benchmarking some representative deep learning based object detection methods on our proposed DIOR dataset in order to provide an overview of the state-of-the-art performance for future research work.

### 5.1 Experimental Setup

In order to guarantee the distributions of training-validation (trainval) data and test data are similar, we randomly selected 11725 remote sensing images (i.e., 50% of the dataset) as trainval set, and the remaining 11738 images are used as test set. The trainval data consists of two parts, the training (train) set and validation (val) set. For each object category and subset, the number of images that contains at least one object instance of that object class is reported in Table 2. Note that one image may contain multiple object classes, so the column totals do not simply equal the sums of each corresponding column. A detection is regarded as correct if its bounding box has more than 50% overlap with the ground truth; otherwise, the detection is seen as a false positive. We conducted all experiments on a computer with a single Intel core i7 CPU, 64 GB of memory, and an NVIDIA Titan X GPU for acceleration.

**Table 2**
Number of images per object class and per subset.

|  | Train | val | Trainval | Test |
|---|---|---|---|---|
| Airplane | 344 | 338 | 682 | 705 |
| Airport | 326 | 327 | 653 | 657 |
| Baseball field | 551 | 577 | 1128 | 1312 |
| Basketball court | 336 | 329 | 665 | 704 |
| Bridge | 379 | 495 | 874 | 1302 |
| Chimney | 202 | 204 | 406 | 448 |
| Dam | 238 | 246 | 484 | 502 |
| Expressway service area | 279 | 281 | 560 | 565 |
| Expressway toll station | 285 | 299 | 584 | 634 |
| Golf course | 216 | 239 | 455 | 491 |
| Ground track field | 536 | 454 | 990 | 1322 |
| Harbor | 328 | 332 | 660 | 814 |
| Overpass | 410 | 510 | 920 | 1099 |
| Ship | 650 | 652 | 1302 | 1400 |
| Stadium | 289 | 292 | 851 | 619 |
| Storage tank | 391 | 384 | 775 | 839 |
| Tennis court | 605 | 630 | 1235 | 1347 |
| Train station | 244 | 249 | 493 | 501 |
| Vehicle | 1556 | 1558 | 3114 | 3306 |
| Wind mill | 404 | 403 | 807 | 809 |
| Total | 5862 | 5863 | 11725 | 11738 |

A total of 12 representative deep learning based object detection methods, which are widely used for object detection in natural scene images and earth observation images, were selected as our benchmark testing algorithms. To be specific, our selections include eight region proposal-based approaches: R-CNN (Girshick et al., 2014), RICNN (with R-CNN framework) (Cheng et al., 2016a), RICAOD (Li et al., 2018), Faster R-CNN (Ren et al., 2017), RIFD-CNN (with Faster R-CNN framework) (Cheng et al., 2019), Faster R-CNN (Ren et al., 2017) with FPN (Lin et al., 2017b), Mask R-CNN (He et al., 2017) with FPN (Lin et al., 2017b) and PANet (Liu et al., 2018b), and four regression-based methods: YOLOv3 (Redmon and Farhadi, 2018), SSD (Liu et al., 2016a), RetinaNet (Lin et al., 2017c), and CornerNet (Law and Deng, 2018). To make fair comparisons, we kept all the experiment settings the same as that depicted in corresponding papers. R-CNN (Girshick et al., 2014), RICNN (Cheng et al., 2016a), RICAOD (Li et al., 2018), and RIFD-CNN (Cheng et al., 2019) are built on the Caffe framework (Jia et al., 2014). Faster R-CNN (Ren et al., 2017), Faster R-CNN (Ren et al., 2017) with FPN (Lin et al., 2017b), Mask R-CNN (He et al., 2017) with FPN (Lin et al., 2017b), PANet (Liu et al., 2018b), RetinaNet (Lin et al., 2017c), and CornerNet (Law and Deng, 2018) are based on the PyTorch re-implementation (Paszke et al., 2017). YOLOv3 uses Darknet-53 framework (Redmon and Farhadi, 2018) and SSD (Liu et al., 2016a) is implemented with TensorFlow (Abadi et al., 2016). Note that, the backbone network is VGG16 model (Simonyan and Zisserman, 2015) for R-CNN (Girshick et al., 2014), RICNN (Cheng et al., 2016a), RICAOD (Li et al., 2018), Faster R-CNN (Ren et al., 2017), RIFD-CNN (Cheng et al., 2019), and SSD (Liu et al., 2016a). YOLOv3 (Redmon and Farhadi, 2018) uses Darknet-53 as its backbone network. For Faster R-CNN (Ren et al., 2017) with FPN (Lin et al., 2017b), Mask R-CNN (He et al., 2017) with FPN (Lin et al., 2017b), PANet (Liu et al., 2018b), and RetinaNet (Lin et al., 2017c), we use ResNet-50 and ResNet-101 (He et al., 2016) as their backbone networks. As regards CornerNet (Law and Deng, 2018), its backbone network is Hourglass-104 (Newell et al., 2016). We used average precision (AP) and mean AP as measures for evaluating the object detection performance. One can refer to (Cheng and Han, 2016) for more details about these two metrics.

**5.2 Experimental Results**

The results of 12 representative methods are shown in Table 3. We have the following observations from Table 3.

(1) The deeper the backbone network is, the stronger the representation capability of the network is and the higher detection accuracy we could obtain. It generally follows the order: ResNet-101 and Hourglass-104 > ResNet50 and Darknet-53 > VGG16. The detection results of RetinaNet (Lin et al., 2017c) with ResNet-101 and PANet (Liu et al., 2018b) with ResNet-101 both achieve the highest mAP of 66.1%. (2) Since CNNs naturally form a feature pyramid through its forward propagation, exploiting inherent pyramidal hierarchy of CNNs to construct feature pyramid networks, such as FPN (Lin et al., 2017b) and PANet (Liu et al., 2018b), could significantly boost the detection accuracy. Using FPN in basic Faster R-CNN and Mask RCNN systems shows great advances for detecting objects with a wide variety of scales. And for this reason, FPN has now become a basic building block of many latest detectors such as RetinaNet (Lin et al., 2017c) and PANet (Liu et al., 2018b). (3) YOLOv3 (Redmon and Farhadi, 2018) could always achieve higher accuracy than other methods for detecting small-sized object instances (e.g., vehicles, storage tanks and ships). Especially for ship class, the detection accuracy of YOLOv3 (Redmon and Farhadi, 2018) achieves 87.40%, which is much better than all other 11 methods. This is probably because that the backbone network of Darknet-53 is specifically designed for object detection task and also the new multi-scale prediction is introduced into YOLOv3, which allows it to extract richer features from three different scales (Lin et al., 2017b). (4) For ship, airplane, basketball court, vehicle, bridge, RIFD-CNN (Cheng et al., 2019), RICAOD (Li et al., 2018) and RICNN (Cheng et al., 2016a) improve the detection accuracies to some extent in comparison with the baseline approaches of Faster R-CNN (Ren et al., 2017) and R-CNN (Girshick et al., 2014). This is mainly because these methods proposed different strategies to enrich feature representations for remote sensing images to address the issue of geospatial object rotation variations. Specifically, RICAOD (Li et al., 2018) designs a rotation-insensitive region proposal network. RICNN (Cheng et al., 2016a) presents a rotation-invariant CNN by adding a new fully-connected layer. RIFD-CNN (Cheng et al., 2019) learns a rotation-invariant and Fisher discriminative CNN by proposing new objective functions yet without changing the CNN model architecture. (5) CornerNet (Law and Deng, 2018) obtains the best results for 9 of 20 object classes, which demonstrates that detecting an object as a pair of bounding box corners is a very promising research direction.

**Table 3**
Detection average precision (%) of 12 representative methods on the proposed DIOR test set. The entries with the best APs for each object category are bold-faced.

| c1 | c2 | c3 | c4 | c5 | c6 | c7 | c8 | c9 | c10 |
|---|---|---|---|---|---|---|---|---|---|
| Airplane | Airport | Baseball field | Basketball court | Bridge | Chimney | Dam | Expressway service area | Expressway toll station | Golf course |
| c11 | c12 | c13 | c14 | c15 | c16 | c17 | c18 | c19 | c20 |
| Ground track field | Harbor | Overpass | Ship | Stadium | Storage tank | Tennis court | Train station | Vehicle | Wind mill |

| | Backbone | c1 | c2 | c3 | c4 | c5 | c6 | c7 | c8 | c9 | c10 | c11 | c12 | c13 | c14 | c15 | c16 | c17 | c18 | c19 | c20 | mAP |
|---|---|---|---|---|---|---|---|---|---|---|---|---|---|---|---|---|---|---|---|---|---|---|
| R-CNN | VGG16 | 35.6 | 43.0 | 53.8 | 62.3 | 15.6 | 53.7 | 33.7 | 50.2 | 33.5 | 50.1 | 49.3 | 39.5 | 30.9 | 9.1 | 60.8 | 18.0 | 54.0 | 36.1 | 9.1 | 16.4 | 37.7 |
| RICNN | VGG16 | 39.1 | 61.0 | 60.1 | 66.3 | 25.3 | 63.3 | 41.1 | 51.7 | 36.6 | 55.9 | 58.9 | 43.5 | 39.0 | 9.1 | 61.1 | 19.1 | 63.5 | 46.1 | 11.4 | 31.5 | 44.2 |
| RICAOD | VGG16 | 42.2 | 69.7 | 62.0 | 79.0 | 27.7 | 68.9 | 50.1 | 60.5 | 49.3 | 64.4 | 65.3 | 42.3 | 46.8 | 11.7 | 53.5 | 24.5 | 70.3 | 53.3 | 20.4 | 56.2 | 50.9 |
| RIFD-CNN | VGG16 | 56.6 | 53.2 | **79.9** | 69.0 | 29.0 | 71.5 | 63.1 | 69.0 | 56.0 | 68.9 | 62.4 | **51.2** | 51.1 | 31.7 | **73.6** | 41.5 | 79.5 | 40.1 | 28.5 | 46.9 | 56.1 |
| Faster R-CNN | VGG16 | 53.6 | 49.3 | 78.8 | 66.2 | 28.0 | 70.9 | 62.3 | 69.0 | 55.2 | 68.0 | 56.9 | 50.2 | 50.1 | 27.7 | 73.0 | 39.8 | 75.2 | 38.6 | 23.6 | 45.4 | 54.1 |
| SSD | VGG16 | 59.5 | 72.7 | 72.4 | 75.7 | 29.7 | 65.8 | 56.6 | 63.5 | 53.1 | 65.3 | 68.6 | 49.4 | 48.1 | 59.2 | 61.0 | 46.6 | 76.3 | 55.1 | 27.4 | 65.7 | 58.6 |
| YOLOv3 | Darknet-53 | **72.2** | 29.2 | 74.0 | 78.6 | 31.2 | 69.7 | 26.9 | 48.6 | 54.4 | 31.1 | 61.1 | 44.9 | 49.7 | **87.4** | 70.6 | **68.7** | **87.3** | 29.4 | **48.3** | 78.7 | 57.1 |
| Faster RCNN with FPN | ResNet-50 | 54.1 | 71.4 | 63.3 | 81.0 | 42.6 | 72.5 | 57.5 | 68.7 | 62.1 | 73.1 | 76.5 | 42.8 | 56.0 | 71.8 | 57.0 | 53.5 | 81.2 | 53.0 | 43.1 | 80.9 | 63.1 |
| | ResNet-101 | 54.0 | 74.5 | 63.3 | 80.7 | 44.8 | 72.5 | 60.0 | 75.6 | 62.3 | 76.0 | 76.8 | 46.4 | 57.2 | 71.8 | 68.3 | 53.8 | 81.1 | 59.5 | 43.1 | 81.2 | 65.1 |
| Mask-RCNN with FPN | ResNet-50 | 53.8 | 72.3 | 63.2 | 81.0 | 38.7 | 72.6 | 55.9 | 71.6 | 67.0 | 73.0 | 75.8 | 44.2 | 56.5 | 71.9 | 58.6 | 53.6 | 81.1 | 54.0 | 43.1 | 81.1 | 63.5 |
| | ResNet-101 | 53.9 | 76.6 | 63.2 | 80.9 | 40.2 | 72.5 | 60.4 | 76.3 | 62.5 | 76.0 | 75.9 | 46.5 | 57.4 | 71.8 | 68.3 | 53.7 | 81.0 | **62.3** | 43.0 | 81.0 | 65.2 |
| RetinaNet | ResNet-50 | 53.7 | 77.3 | 69.0 | 81.3 | 44.1 | 72.3 | 62.5 | 76.2 | 66.0 | 77.7 | 74.2 | 50.7 | 59.6 | 71.2 | 69.3 | 44.8 | 81.3 | 54.2 | 45.1 | 83.4 | 65.7 |
| | ResNet-101 | 53.3 | 77.0 | 69.3 | **85.0** | 44.1 | 73.2 | 62.4 | 78.6 | 62.8 | 78.6 | 76.6 | 49.9 | 59.6 | 71.1 | 68.4 | 45.8 | 81.3 | 55.2 | 44.4 | 85.5 | **66.1** |
| PANet | ResNet-50 | 61.9 | 70.4 | 71.0 | 80.4 | 38.9 | 72.5 | 56.6 | 68.4 | 60.0 | 69.0 | 74.6 | 41.6 | 55.8 | 71.7 | 72.9 | 62.3 | 81.2 | 54.6 | 48.2 | **86.7** | 63.8 |
| | ResNet-101 | 60.2 | 72.0 | 70.6 | 80.5 | 43.6 | 72.3 | 61.4 | 72.1 | 66.7 | 72.0 | 73.4 | 45.3 | 56.9 | 71.7 | 70.4 | 62.0 | 80.9 | 57.0 | 47.2 | 84.5 | **66.1** |
| CornerNet | Hourglass-104 | 58.8 | **84.2** | 72.0 | 80.8 | **46.4** | **75.3** | **64.3** | **81.6** | **76.3** | **79.5** | **79.5** | 26.1 | **60.6** | 37.6 | 70.7 | 45.2 | 84.0 | 57.1 | 43.0 | 75.9 | 64.9 |

While the results on some object categories are promising, there exists substantial improvement space for almost all object categories. For some object classes, e.g., bridge, harbor, overpass, and vehicle, the detection accuracies are still very low, and the currently existing methods are difficult to obtain satisfactory results. This probably attributes to the relatively low image quality and the complex and cluttered background in aerial images, when compared with natural scene images. This also indicates that the proposed DIOR dataset is a challenging benchmark for geospatial object detection. In the future work, some novel training scheme including SNIP (Singh and Davis, 2018) and SNIPER (Singh et al., 2018b) can be applied to many existing detectors, such as Faster R-CNN (Ren et al., 2017), Mask R-CNN (He et al., 2017), R-FCN (Dai et al., 2016), deformable R-FCN (Dai et al., 2017), and so on, to obtain better results.

## 6. Conclusions

This paper first highlighted the recent progress of object detection, including benchmark datasets and the state-of-the-art deep learning-based approaches, in both the computer vision and earth observation communities. Then, a large-scale and publicly available object detection benchmark dataset is proposed. This new dataset can help the earth observation community to further explore and validate deep learning based methods. Finally, the performances of some representative object detection methods are evaluated by using the proposed dataset and the experimental results can be regarded as a useful performance baseline for future research.

## Acknowledgements

This work was supported in part by the National Natural Science Foundation of China under Grant 61573284, Grant 61772425, and Grant 61790552, in part by the Project of Science and Technology Innovation of Henan Province under Grant 142101510005, in part by the Young Star of Science and Technology in Shaanxi Province under grant 2018KJXX-029, and in part by the Aerospace Science Foundation of China under Grant 2017ZC53032.